\documentclass[a4paper,fleqn]{cas-dc}

\usepackage{hyperref}

\usepackage{caption}
\usepackage{subcaption}
\usepackage{multirow}
\usepackage{graphicx}
\usepackage{multicol}
\usepackage{array}
\usepackage{comment}
\usepackage{longtable}
\usepackage{amsmath,amsfonts}
\usepackage{textcomp}
\usepackage{xcolor}
\usepackage{amsthm}
\usepackage{ulem, soul}
\usepackage{algorithm}
\usepackage{algorithmic}
\usepackage[numbers]{natbib}
\usepackage[capitalize,noabbrev]{cleveref}
 \def\BibTeX{{\rm B\kern-.05em{\sc i\kern-.025em b}\kern-.08em
    T\kern-.1667em\lower.7ex\hbox{E}\kern-.125emX}}
\usepackage{balance}

\DeclareMathOperator*{\argmax}{arg\,max}
\DeclareMathOperator*{\argmin}{arg\,min}
\theoremstyle{definition}
\newtheorem{definition}{Definition}[section]
\newcolumntype{P}[1]{>{\centering\arraybackslash}p{#1}}

\def\tsc#1{\csdef{#1}{\textsc{\lowercase{#1}}\xspace}}
\tsc{WGM}
\tsc{QE}
\tsc{EP}
\tsc{PMS}
\tsc{BEC}
\tsc{DE}

\begin{document}

\shorttitle{Nonlinear Transformations Against Unlearnable Datasets}
\shortauthors{T. Hapuarachchi et~al.}
\title{Nonlinear Transformations Against Unlearnable Datasets}


\author[1]{Thushari Hapuarachchi}[orcid=0009-0001-0067-6997]
\ead{saumya2@usf.edu}


\author[1]{Jing Lin}
\ead{jinglin314@gmail.com}

\author[1]{Kaiqi Xiong}[orcid=0000-0003-2933-8083]
\ead{xiongk@usf.edu}
\cormark[1]


\author[2]{Mohamed Rahouti}
\ead{mrahouti@fordham.edu}

\author[1]{Gitte Ost}
\ead{gitteost@usf.edu}

\affiliation[1]{organization={University of South Florida},
                city={Tampa},
                state={FL},
                country={USA}}

\affiliation[2]{organization={Fordham University},
                city={New York}, state={NY},
                country={USA}}

\cortext[cor1]{Corresponding author}


\begin{abstract}
Automated scraping stands out as a common method for collecting data in deep learning models without the authorization of data owners. Recent studies have begun to tackle the privacy concerns associated with this data collection method. Notable approaches include Deepconfuse, error-minimizing, error-maximizing (also known as adversarial poisoning), Neural Tangent Generalization Attack, synthetic, autoregressive, One-Pixel Shortcut, Self-Ensemble Protection, Entangled Features, Robust Error-Minimizing, Hypocritical, TensorClog, and Provably Unlearnable Examples. The data generated by those approaches, called ``unlearnable" examples, are prevented ``learning" by deep learning models. In this research, we investigate and devise an effective nonlinear transformation framework and conduct extensive experiments to demonstrate that a deep neural network can effectively learn from the data/examples traditionally considered unlearnable produced by the above 13 approaches. The resulting approach improves the ability to break unlearnable data compared to the linear separable technique recently proposed by researchers. Specifically, our extensive experiments show that the improvement ranges from 0.65\% to 234.74\% for the unlearnable CIFAR10 datasets generated by those 13 data protection approaches, except the One-Pixel Shortcut and Provably Unlearnable Examples. Moreover, the proposed framework achieves over 100\% improvement in test accuracy for Autoregressive and REM approaches compared to the linear separable technique. Our findings suggest that these approaches are inadequate in preventing unauthorized uses of data in machine learning models. There is an urgent need to develop more robust protection mechanisms that effectively thwart an attacker from accessing data without proper authorization from the owners.
\end{abstract}


\maketitle



\section{Introduction}

Deep learning typically requires a large dataset to achieve reliable performance, prompting researchers to make significant efforts in scraping data from the Internet. However, the owners of datasets may have a serious concern about the unauthorized use of the databases, including copyright infringement and privacy violations, especially in domains like media streaming and privacy-preserving applications~\cite{DBLP:conf/sp/ShokriSSS17}. The infringement and violations have inspired a variety of research studies to avoid such a violation of data collection. Among these, a generalization attack stands out as a prominent method for impeding a deep neural network (DNN) model from effectively learning from a provided dataset. It is a type of data poisoning attack wherein a specific portion of training data (e.g., a portion of data the owners aims to safeguard from unauthorized use) is altered, hindering the learning process and leading to a deficiency of generalization manifested as poor model accuracy on unseen data. It must be pointed out that the significance of data perturbation must be {\it minor} so that legitimate users can still use the dataset. The crafted data are called {\it unlearnable datasets} or {\it unlearnable examples}.

In this research, we investigate 13 well-known approaches to preventing datasets from unauthorized uses by deep learning models. They are Deepconfuse~\cite{DBLP:conf/nips/FengCZ19}, error-minimizing~\cite{DBLP:conf/iclr/HuangME0021}, error-maximizing (also known as adversarial poisoning)~\cite{DBLP:journals/corr/abs-2106-10807}, Neural Tangent Generalization Attack (NTGA)~\cite{yuan2021neural}, synthetic~\cite{yu2021indiscriminate}, autoregressive~\cite{DBLP:journals/corr/abs-2206-03693}, One-Pixel Shortcut (OPS)~\cite{DBLP:conf/iclr/WuCXH23}, Self-Ensemble Protection (SEP)~\cite{chen2022self}, Entangled Features (EntF)~\cite{DBLP:conf/iclr/00020000023}, Robust Error-Minimizing (REM)~\cite{DBLP:conf/iclr/FuHLST22}, Hypocritical~\cite{DBLP:conf/nips/TaoFYHC21}, TensorClog~\cite{DBLP:journals/access/ShenZM19}, and Provably Unlearnable Examples (PUE)~\cite{wang2024provably} approaches. These approaches can be formulated as a bi-level optimization, which is usually very difficult to be solved efficiently unless the learning model is convex \cite{koh2017understanding, jagielski2018manipulating}. For instance, the error-minimizing approach makes personal data completely unusable by solving a min-min optimization problem, in which an iterative process is developed to minimize the training loss with respect to the $L_p$-norm bounded noise and model weights, respectively.

As pointed out {by}~\cite{yu2021indiscriminate}, unlearnable perturbations can effectively disrupt DNN training due to linear separability. However, not all existing unlearnable perturbations exhibit this characteristic as revealed in~\cite{DBLP:journals/corr/abs-2305-19254}. Additionally, the authors present an attack on unlearnable data leveraging the linear separability inherent in such perturbations, referred to as the \textit{orthogonal projection attack} (OPA). Nevertheless, their findings suggest that the effectiveness of OPA diminishes when applied to nonlinear perturbations like autoregressive ones in \cite{DBLP:journals/corr/abs-2206-03693}.

In this paper, we explore 13 advanced data protection approaches. Our main contributions include:

\begin{itemize}
    \item Propose an effective nonlinear transformation framework designed to circumvent data protection measures, thereby facilitating the training of DNNs on augmented, previously deemed unlearnable data. The proposed framework applies to DNN models for image classification tasks.
    \item The proposed nonlinear transformation framework improves the linear separable technique given in~\cite{DBLP:journals/corr/abs-2305-19254} for all 13 data protection approaches, except for OPS in~\cite{DBLP:conf/iclr/WuCXH23} and PUE in~\cite{wang2024provably}. The improvement is very significant for the six approaches: NTGA, Deepconfuse, Error-minimizing, Error-maximizing, Autoregressive, and REM. In particular, the proposed framework achieves over 100\% improvement in test accuracy for Autoregressive and REM compared to the linear separable technique.
    \item Demonstrate through extensive experiments that the nonlinear transformation technique diminishes the efficacy of data protection strategies, empowering DNNs to acquire knowledge from previously deemed 'unlearnable' data with an accuracy comparable to training on pristine data. This underscores the shortcomings of existing data protection methods.
    \item Illustrate experimentally that data protection methods can be circumvented by incorporating clean data from external sources (or partially perturbed data). This underscores the necessity for future data protection strategies to address and mitigate these vulnerabilities.
\end{itemize}

The rest of this paper is organized as follows: we first introduce the background on clean-label generalization attacks and summarize the key "unlearnable example" protection methods considered, including the OPA baseline; we then formalize the attacker’s objective and problem formulation, followed by a detailed description of our nonlinear-transformation framework (threat model, transformation/model selection loop, and the specific transformations used). Next, we present an extensive experimental evaluation, primarily on CIFAR-10 across 13 protection approaches, benchmarking against OPA and other relevant baselines, and we then broaden the analysis with additional experiments and discussion (including results on other datasets such as MNIST/ImageNet and additional protection methods) before concluding with the main takeaways and implications for designing more robust data-protection mechanisms.

\section{Preliminaries}
Various data poisoning attacks can target machine learning algorithms, with our specific emphasis here on generalization attacks. In the context of a generalization attack, adversaries endeavor to manipulate the dataset, disrupting the training process of the DNN model. The ultimate goal is to yield a model with compromised generalizability or diminished capacity for generalization.

Generalization attacks on machine learning models is contingent upon the adversary's ability to manipulate the training data. They are broadly classified into two categories: dirty-label attacks and clean-label attacks. This study primarily focuses on clean-label-based generalization attacks, as a substantial portion of web data is typically unlabeled before the data collection process.

\begin{definition}[Clean-label generalization attack] 
Let $D=(X_D, Y_D)$ be a training set, where $X_D \in \mathbb{R}^{n \times d}$ is a set of training images, $n \in \mathbb{N}$ is the number of training images,  and $d \in \mathbb{N}$ is the dimension of training images;  $Y_D \in \mathbb{R}^{n \times c}$ is a set of training outputs and $c \in \mathbb{N}$ is the dimension of training labels. Similarly, we denote a test set by $V= (X_V,  Y_V)$, where $m \in \mathbb{N}$ is the number of test images,  $X_V \in \mathbb{R}^{m \times d}$ is a set of test images, and  $Y_V \in \mathbb{R}^{m \times c}$ is a set of test labels. Let $f(.; \theta)$ be a machine learning model parameterized by $\theta$. Then, the generalization attack generates the ``unlearnable examples" by solving the following bi-level optimization problem~\cite{yuan2021neural}:
\begin{equation} \label{eq:eq1}
    \argmax_{ \|g_{\xi}(X_D)\|_{p} \le \epsilon}\mathcal{L}_V( f(X_V; \theta^*), Y_V)  
\end{equation}
$$
\hspace{-3mm}\mbox{subject to}\hspace{3mm}  \theta^* \in \argmin_{\theta}\mathcal{L}_D( f(X_D + g_{\xi}(X_D); \theta), Y_D ),
$$
where $\mathcal{L}_V$ and $\mathcal{L}_D$ are the loss functions of test and training sets, respectively; $g_{\xi}$ is a noise generator characterized by the weight parameter $\xi$, and $\epsilon$ represents the maximum allowable perturbation or noise specified by a user.
Since $g_{\xi}$ is only added to $X_D$ in (\ref{eq:eq1}) and the label $Y_D$ is not modified, it is called {\it a clean-label generalization attack}. 
\end{definition}

A trivial solution to the bi-level optimization problem (\ref{eq:eq1}) is to alternatively update $\theta^*$ over poisoned data $X_D + g_{\xi}(X_D)$ by using the gradient {\it descent} method and update $g_{\xi}$ over clean test data $X_V$ by using the gradient {\it ascent} method. However, achieving convergence of both weight parameters $\theta^*$ and $g_{\xi}$ is intractable in practice. Over the past few years, various data protection approaches have been introduced to solve this bi-level optimization problem.

\subsection{Data Protection Approaches}\label{sec:Dataprotection}
We introduce several well-known data protection approaches.

\textbf{Deepconfuse:} \cite{DBLP:conf/nips/FengCZ19} proposed the Deepconfuse approach to solving a simpler version of the bi-level optimization problem (\ref{eq:eq1}). They relaxed the constraint in (\ref{eq:eq1}) by decoupling the alternating update procedure for stability and memory efficiency to avoid the storage of the gradient update of $\theta_i$ and model $g_{\xi}$ as an auto-encoder. Their objective is to find a noise generator $g_{\xi^*}$ that results in a classifier, $f$, with the worst test accuracy.

\textbf{Error-minimizing:}
This approach in~\cite{DBLP:conf/iclr/HuangME0021} makes data unlearnable for a deep learning model by minimizing the training loss. 
The model can no longer learn anything from these examples since the training loss is close to zero. 
Hence, this approach protects against the unauthorized exploitation of the data. The following min-min bi-level optimization problem generates error-minimizing noise $\delta$ to inject into clean training input $D$ in order to make $D$ unusable for DNNs \cite{DBLP:conf/iclr/HuangME0021}:
\begin{equation} \label{eq2}
\min_{\theta} \left[ \min_\delta \mathcal{L}_{D}(f(X_D+\delta; \theta),Y_D) \right],
\end{equation}
subject to \[\|\delta\|_p \le \epsilon,\]
where $\mathbf{\delta} = [ \delta_1, \delta_2, ..., \delta_n]$ is the perturbation. Both the noise $\delta$ and the weight parameter $\theta$ are found by minimizing the classification loss $\mathcal{L}_{D}$. $x'_{i} = x_{i} + \delta_{i}$ is the $i$-th unlearnable example. According to 
\cite{DBLP:conf/iclr/HuangME0021}, this type of noise is called {\it sample-wise noise} since the noise is generated separately for each example. They also proposed {\it class-wise noise}, where all examples in the same class have the same noise.

To solve this min-min bi-level optimization problem~(\ref{eq2}), they proposed an iterative algorithm by repeatedly performing $M$ steps of optimization for $\theta$ (this is the regular model training), followed by calculating $\delta$ over $D$ based on the Projected Gradient Descent (PGD) in \cite{DBLP:conf/iclr/MadryMSTV18}. The iterative process stops once 
the error rate falls below the threshold defined by the user-specified parameter $\lambda$.

\textbf{Error-maximizing:} 
\cite{DBLP:journals/corr/abs-2106-10807} decided not to solve the general bi-level problem~(\ref{eq:eq1}) but instead solved the following empirical loss maximizing problem:
\begin{equation}
\label{eq6}
\max_{\|\delta\|_{p} \le \epsilon} \Big[ \mathcal{L}_{D}(f(X_D+\delta;\theta^*),Y_D) \Big],
\end{equation}
where $\theta^*$ denotes the parameters of a model trained on clean data \cite{DBLP:journals/corr/abs-2106-10807}. Most attacks in \cite{DBLP:journals/corr/abs-2106-10807} are bounded by $l_\infty$-norm with $\epsilon = 8/255$. The optimization problem (\ref{eq6}) is solved with 250 steps of PGD. \cite{DBLP:journals/corr/abs-2106-10807} also used differentiable data augmentation when crafting the poisons. 

\cite{DBLP:journals/corr/abs-2106-10807} further introduced a variant of (\ref{eq6}), called the class targeted adversarial attack: 
\begin{equation}
\label{eq7}
\max_{\|\delta\|_{p} \le \epsilon} \Big[ \mathcal{L}_{2}(f(x_i+\delta_i;\theta^*),g(y_i)) \Big],
\end{equation}
where $g$ is a permutation on the label space. For crafting class targeted attacks, they labeled $i \rightarrow i +3$ for CIFAR-10 \cite{DBLP:journals/corr/abs-2106-10807}.

\textbf{NTGA:} 
Before describing NTGA, we first review the Neural Tangent Kernel (NTK), which was introduced by~\cite{jacot2018neural}. NTK is a kernel describing the DNN evolution during the training by gradient descent. NTK becomes a constant in the infinite-width limit for most common neural network models (i.e., architectures) and enables the examination of neural network models through kernel methods-based theoretical tools. 

Using the Gaussian process $\bar{f}$ with a deterministic kernel to approximate a class of wide neural networks,
~\cite{yuan2021neural} simplified the bi-level optimization problem in (\ref{eq:eq1}) as:
\begin{equation} \label{eq:eq6}
    \argmax_{ \|g_{\xi}(X_D)\|_{p} \le \epsilon}\mathcal{L}_V( \bar{f}(X_V,X_D, g_{\xi}(X_D), Y_D, t), Y_V),
\end{equation}
where $t$ is the time step at which an attack takes effect during training.
This eliminates the need to find the weight parameter $\theta$ or know the model architecture. This optimization problem can be easily solved with the projected gradient ascent without iterating through the training steps, as in Deepconfuse attacks \cite{DBLP:conf/nips/FengCZ19}. 

\textbf{Synthetic:}
Observing that the advanced techniques described above generate almost linear separable perturbations, 
\cite{yu2021indiscriminate} developed a two-stage process to protect the data. First, they randomly generated some normally distributed noise $\eta$, for some integer $k$ such that $s^2=k*p^2$, where $s \times s$ is the image dimension and $p \times p$ is patch dimension \cite{yu2021indiscriminate}. Then, they cut that image into $k$ patches, where each element in patch $i$ has the same value, which is the $i$-th element of $\eta$. These patches together consist of synthetic noise for that image.

\textbf{Autoregressive:} 
~\cite{DBLP:journals/corr/abs-2206-03693} crafted perturbations using autoregressive (AR) processes, resulting in unlearnable data resistance to common defenses such as adversarial training and ``strong" data augmentations; e.g., CutMix, Cutout, and Mixup~\cite{DBLP:conf/iccv/YunHCOYC19}. Unlike error-minimizing and error-maximizing noise, AR perturbations do not involve a surrogate model; hence, they are faster to generate. AR perturbations are crafted by using the linear dependence on neighboring pixels. Equation (\ref{eq:AR_model}) represents an AR process based on $p$ past observations, denoted by $AR (p)$. 
It forms a filter with a size of $(p + 1)$ using elements $\beta_p, \dots, \beta_1$, and assigns a value of $-1$ to the $(p+1)^{th}$ entry of the filter. 
\cite{DBLP:journals/corr/abs-2206-03693} refer to this filter as an AR filter: 
\begin{equation}
\label{eq:AR_model}
    x_t=\beta_1 x_{t-1}+\beta_2x_{t-2}+\dots+\beta_{p-1} x_{t-p-1}+\beta_p x_{t-p}+\epsilon_t
\end{equation}

\textbf{Other approaches:} In addition to these approaches, there are other ways to generate unlearnable data. Notably, 
\cite{DBLP:conf/iclr/FuHLST22} proposed an extended version of training examples' error reduction called {\it robust error-minimizing noise}. In contrast to error-minimizing noise, robust error-minimizing noise provides defense against adversarial training. Moreover, \cite{DBLP:journals/corr/abs-2111-10130} crafted a noise called ADVersarially Inducing Noise (ADVIN) to make data unlearnable using robust features resistant to adversarial training. After showing that error-maximizing noise is ineffective against unsupervised contrastive learning models, 
\cite{DBLP:journals/corr/abs-2202-11202} introduced a novel data protection approach against contrastive learning models. Further, \cite{sadasivan2023fun} proposed a filter-based poisoning attack using convolutional filters that can craft successful unlearnable datasets. Recently,~\cite{DBLP:journals/corr/abs-2205-12141} studied unlearnable examples and proposed the One-Pixel Shortcut attack, a model-free technique to generate unlearnable samples. They modified a single pixel from every image, which fools DNN models during training. CUDA in~\cite{sadasivan2023cuda} is another recently proposed method to protect data from unauthorized use. It adds protection by blurring images using randomly generated class-wise convolutional filters. Moreover, the study by Gong et al.~\cite{gong2025armor} proposed a data protection approach that is robust against augmentation techniques such as Fast AutoAugment. The study by Zhang et al.~\cite{zhang2023unlearnable} proposed a novel data protection approach that does not rely on the target classes of images but instead depends on generated clusters within the dataset. Consequently, they focused on diverse real-world datasets that can be clustered effectively, rather than standard benchmark datasets like CIFAR-10 and MNIST.

\subsection{Orthogonal Projection Attack (OPA)}

\cite{DBLP:journals/corr/abs-2305-19254} proposed an attack against the data protection approaches discussed in section~\ref{sec:Dataprotection}. They challenged the  notion that unlearnable perturbations must exhibit linear separability across classes for effective exploitation in~\cite{yu2021indiscriminate}. They  demonstrated it using a counter example, autoregressive perturbations, which defy linear separability. However, OPA relies on linear separability to break unlearnable perturbations.

Initially,~\cite{DBLP:journals/corr/abs-2305-19254} trained a linear logistic regression model on the unlearnable dataset to capture linear features in the data. Then, they performed QR decomposition on the obtained feature matrix. The resulting Q matrix can be considered the orthonormal basis of the captured linear space. Subsequently, unlearnable images are orthogonally projected into this space, effectively removing the linear features from the images. This results in the recovery of the unlearnable images.

Additionally, they demonstrated that their approach is more effective against class-wise linearly separable perturbations, such as OPS in \cite{DBLP:journals/corr/abs-2205-12141}, and synthetic examples in~\cite{yu2021indiscriminate} but less effective against nonlinear perturbations, such as autoregressive. Hence, our study intends to employ nonlinear transformations to break such complex unlearnable perturbations.

\section{Problem Statement and Formulation}

\subsection{Motivation}

The goal of this research is to evaluate and systematically defeat unlearnable data protection techniques that aim to prevent unauthorized training of machine learning models on shared datasets. These techniques generate adversarial perturbations that are intentionally crafted to preserve visual semantics for human viewers while rendering the data ``unlearnable" to machine learning models. Despite their increasing adoption, it remains unclear how resilient these defenses are to realistic preprocessing operations or architectural variations during model training.

Unlearnable example generation methods are typically formulated as bi-level optimization problems. The inner loop trains a model on perturbed data to minimize loss, while the outer loop maximizes the loss on a clean test set—producing perturbed training examples that degrade generalization. However, solving this bi-level optimization efficiently depends on model access and assumptions such as linear separability or kernel-based approximations (e.g., NTGA). The resulting defenses, although diverse in formulation (e.g., Deepconfuse, REM, EntF), often rely on introducing perturbations that are model-specific, label-preserving, and structure-aware.

Our core problem is to determine whether these defenses can be bypassed using only publicly available tools and pretrained models, without access to the protection algorithms themselves. In particular, we investigate whether nonlinear image transformations (that are applied in a model-adaptive and feedback-driven manner) can restore learnability in datasets previously made unlearnable.

Thus, our objective is to break unlearnable datasets using black-box nonlinear transformation pipelines guided by test accuracy, thereby challenging the robustness and generalization claims of current data protection mechanisms. The problem is significant given the growing reliance on open-source datasets and the increasing availability of pre-trained deep models that adversaries may exploit.

\subsection{Problem Formulation}

As the attacker, our goal is to reduce the impact of unlearnable datasets using nonlinear transformations.
This endeavor can be conceptualized as solving the following optimization problem:
\begin{equation} \label{eq:eq2}
   \argmin_{A} \max_{ \|g_{\xi}(X_D)\|_{p} \le \epsilon}\mathcal{L}_V( f(X_V; \theta^*), Y_V)  
\end{equation} 
\begin{equation*}
\hspace{-1mm}\mbox{subject to}\hspace{1mm}    \theta^*\in \argmin_{\theta}\mathcal{L}_D( f(A(X_D + g_{\xi}(X_D)); \theta), Y_D ),
\end{equation*}
where $\mathcal{A}$ represents a set of nonlinear transformations and $A$ is a vector of those transformations whose $i$-th element is $a_i\in \mathcal{A}$ and represents the data augmentation technique applies to $i$-th image in $X_D+ g_{\xi}(X_D)$. The goal of nonlinear transformations is to expand a training set by using class-preserving conversions such as threshold binary~\cite{dao2019kernel, qiu2021deepsweep}.

\section{Methodology} \label{sec:methodology}
Data manipulation has become a pervasive technique across diverse research problems, yet leveraging it effectively to address various challenges remains a formidable and intricate task. This research meticulously examines the characteristics of each nonlinear transformation, identifying specific methods employed to address challenges of breaking unlearnable datasets.

\subsection{Threat Model}

We consider a setting where a \textbf{data owner (defender)} seeks to publicly share a dataset $X_D$ (e.g., on the Internet) while preventing unauthorized machine learning models from extracting meaningful patterns of the dataset. To achieve this goal, the defender applies a perturbation $\delta$, yielding a perturbed dataset $X_U = X_D + \delta$, commonly referred to as an \textit{unlearnable dataset}. The perturbation $\delta$ is bounded by $||\delta||_p \leq \epsilon$, where $\epsilon$ is the maximum allowable perturbation to maintain imperceptibility. The perturbation is crafted using one of 13 data protection techniques to ensure that any model trained on $X_U$ fails to generalize to clean, unprotected test data $X_V$.

This study explores the vulnerabilities of these twelve unlearnable datasets by playing the role of the attacker. The \textbf{adversary (attacker)} is an entity with access to both the publicly released perturbed dataset $X_U$ and a clean test set $X_V$. The dataset $X_v$ is a much smaller, partial dataset compared to the unlearnable dataset $X_U$. Since the perturbations are both bounded and imperceptible, the attacker can still perceive the content of images in $X_U$. As a result, the attacker can find clean images matching the target classes or visual features in the unlearnable dataset—either from public platforms or their own dataset. The attacker does not involve the test set $X_V$ during the training process. The attacker can only use it to evaluate the performance of the attack in terms of test accuracy. The attacker's goal is to recover the learnability of $X_U$. That is, to train a model $f$ on $X_U$ that generalizes effectively to $X_V$, thereby breaking the protection of the dataset applied by the defender.

We our summarize our key assumptions below.
\begin{itemize}
    \item The attacker has \textbf{no knowledge} of the defender's model architecture or learning strategy.
    \item The perturbations added by the defender are \textbf{label-preserving} and \textbf{visually imperceptible} such that the resulting data maintains human semantic interpretability.
    \item The attacker has \textbf{black-box access} to pretrained models and publicly available tools (e.g., OpenCV, Keras), which may be adapted to build the learning function $f$.
    \item The attacker can \textbf{use} $X_V$, which is unprotected and public, exclusively for evaluating the performance of the proposed framework and cannot use $X_V$ in the training process.
    \item The attacker can apply \textbf{any combination of nonlinear transformations} to $X_U$ in a model-agnostic manner to restore its learnability.
\end{itemize}

This adversarial interaction captures a realistic scenario in which a malicious learner attempts to bypass data obfuscation defenses using open-source infrastructure and unlabeled auxiliary datasets. Our focus is on evaluating the resilience of current protection techniques under this attack model and determining whether nonlinear preprocessing pipelines can defeat state-of-the-art unlearnable dataset generators.

\subsection{The Proposed Framework}

\begin{figure}
    \centering
    \includegraphics[scale=0.85]{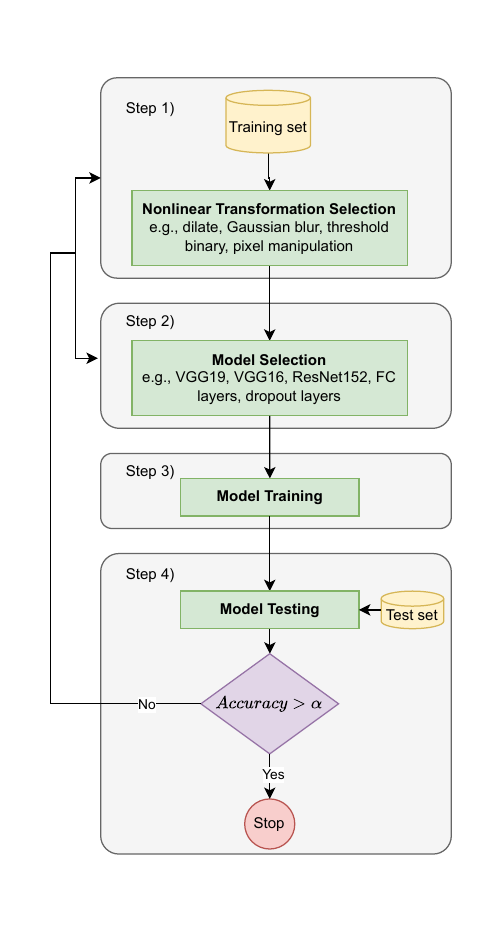}
    \caption{The proposed framework consisting of Nonlinear transformations identification, Model Selection, Model Training,  and Model Testing. $\alpha$ is the expected or predefined accuracy.}
    \label{fig:flowchart1}
\end{figure}

To systematically break unlearnable datasets generated by existing data protection approaches, we propose a framework composed of four key components, executed sequentially:
\begin{enumerate}
    \item \textbf{Nonlinear Transformation Identification:} Apply one or more nonlinear transformations to the unlearnable dataset. Our primary utilization of nonlinear transformations relies on the OpenCV Library, a python package for computer vision. Those nonlinear transformations include dilate~\cite{gonzalez2008digital}, Gaussian blur~\cite{DBLP:journals/tsp/ChenM09}, erode~\cite{DBLP:journals/pami/HaralickSZ87}, threshold binary~\cite{komarudin2015designing}, threshold binary inverse~\cite{DBLP:conf/iccsa/GervasiCM13}, and pixel manipulation~\cite{chaumette2012image}. Additionally, for rotation, horizontal flipping, and other transformations, we employ the Keras ImageDataGenerator~\cite{rahmatullah2021effectiveness}. These transformations effectively augment the dataset size for training purposes. We refer to this augmented and transformed dataset as the \textit{learnable training set}. We further explain how to obtain the learnable training set in Section~\ref{Sec:Data_augmentation}.
    \item \textbf{Model Selection:} Choose a base neural network (e.g., VGG16, VGG19, ResNet152), optionally modifying its architecture (e.g., adding fully connected and dropout layers). While using a pretrained model is not mandatory, it is commonly convenient. In our experiments, a pretrained model is employed in all cases except for the unlearnable MNIST experiment.
    \item \textbf{Model Training:} Train the selected model using the transformed (augmented) learnable dataset obtained in Step 1). It is important to note that there is no clean data involved in this training phase. The model learns only from the unlearnable dataset transformed by nonlinear transformations, which we call the \textit{learnable training set}. 
    \item \textbf{Model Testing:} Evaluate the trained model on a clean (unperturbed) test set to monitor underfitting/overfitting and trigger model adjustments as needed. With a test accuracy surpassing a threshold value $\alpha$, we stop the process. If the test accuracy falls below expectations, we may explore alternative nonlinear transformations on the training set or consider increasing model complexity (e.g., transitioning from VGG16 to VGG19) to bolster learning capabilities. For instance, we can make adjustments to the model architecture, learning rate, batch size, and number of epochs. Additionally, exploring different pretrained models is an alternative option.
\end{enumerate}
These components are also illustrated in Fig. \ref{fig:flowchart1} and detailed in the following subsections. The framework is designed to generalize across various unlearnable data types and model architectures, and includes heuristics (Algorithm \ref{alg:Alg1}) for selecting effective nonlinear transformations based on empirical test accuracy.

\begin{figure*}[ht]
    \centering
    \includegraphics[scale=0.8]
    {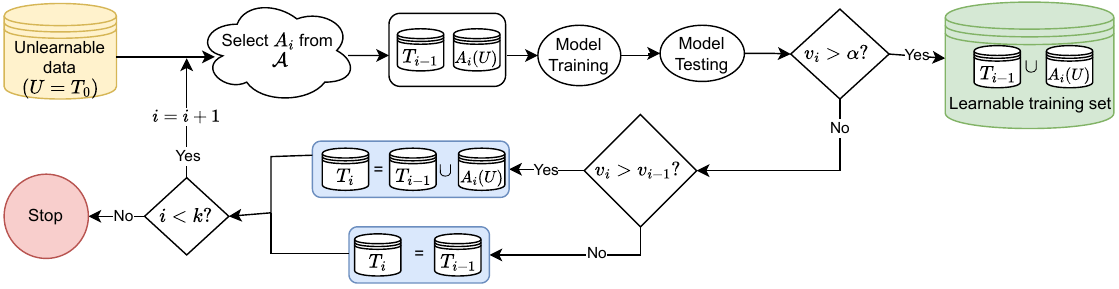}
    \caption{Graphical illustration of the proposed procedure for breaking an unlearnable dataset. In each iteration $i$ ranging from $1$ to $k$, a nonlinear transformation, $A_i$, is applied to the unlearnable dataset ($U$) from the collection ($\mathcal{A}$), denoted as $A_i(U)$.
    The model is then trained on the augmented dataset, and testing is carried out on a clean test dataset, which is much smaller than the unlearnable training dataset. If there's an improvement in test accuracy compared to the previous iteration, the augmented dataset is incorporated into the training set. Here, $v_i$ represents the test accuracy in the $i^{th}$ iteration, with $v_0$ being the test accuracy of the model trained on the initial unlearnable dataset.}
    \label{fig:flowchart2}
\end{figure*}

\begin{algorithm}[ht]
\caption{Generating A Learnable Training Set}
\begin{algorithmic}[1]
    \STATE {\bfseries Input:} Unlearnable training set $(U)$, Clean test set, Pretrained model $(M)$, Number of iterations $(k)$, Space of nonlinear transformations: $\mathcal{A}=\{A_1,A_2, \dots \}$, and Target accuracy $(\alpha)$
    \STATE {\bfseries Output:} Augmented learnable training set $(L)$
    \STATE $L \leftarrow U$
    \STATE Train $M$ on $L$ and obtain test accuracy $v_0$
    \FOR{$i=1$ {\bfseries to} $k$}
        \STATE $T_i \leftarrow L + A_i(U)$ \COMMENT{
        The augmented training set $T_i$} 
        \STATE Train $M$ on $T_i$ and obtain test accuracy $v_i$
        \IF{$v_i > \alpha$}
            \STATE $L \leftarrow T_i$\
            \STATE {\bfseries Break} the loop
        \ELSIF{$v_{i-1} < v_i$}
            \STATE $L \leftarrow T_i$\
        \ELSE
            \STATE $L \leftarrow L$\
        \ENDIF
    \ENDFOR
\end{algorithmic}
\label{alg:Alg1}
\end{algorithm}

\subsection{Breaking Unlearnable Datasets}\label{Sec:Data_augmentation}
The pivotal stage in the proposed framework involves identifying suitable nonlinear transformations to address unlearnable datasets, a task characterized by its challenges and time-intensive nature. 

Solving the optimization problem in (\ref{eq:eq2}) is intractable, so we present a heuristic approach (Algorithm~\ref{alg:Alg1}) to find a set of proper nonlinear transformations. This algorithm provides a procedure for discovering nonlinear transformations to break an unlearnable dataset. Its time complexity is $k$ times that of the sum of the training and testing times for the model $M$. Fig.~\ref{fig:flowchart2} graphically illustrates this procedure. 

Our initial step involves choosing a {nonlinear transformation from the spectrum available. Then, we systematically expand the training set by applying each transformation sequentially. It is pivotal to visually inspect a sample of augmented data at each step, as not all techniques yield meaningful images for every dataset. For instance, Threshold Binary and Threshold Binary Inverse transformations may not generate meaningful images for the CIFAR-10 dataset.

Following the application of each technique, we assess the model's performance by obtaining test accuracy. The test data remain unperturbed (clean). If there is an improvement in test accuracy, we retain the expanded dataset for the subsequent iteration. The process continues until the model achieves the target accuracy ($\alpha$), at which point we conclude the dataset expansion.

In implementing the aforementioned method, we predominantly employed conventional nonlinear transformations, encompassing threshold binary, threshold binary inverse, color channel manipulation, erode, dilate, and Gaussian blur. Threshold binary and threshold binary inverse are commonly applied to grayscale images but can also be used in color images to delineate the primary object from its background. Color channel manipulation is akin to grayscale transformation, involving alterations to the values of one or more color channels in diverse ways.

Although data augmentation methods have been used in different applications, there are several key challenges in applying them against unlearnable data. These challenges include: (1) careful identification of appropriate data augmentation methods, (2) the potential requirement for multiple data augmentation methods to mitigate the effects of unlearnable perturbations, and (3) high computational cost due to the use of multiple data augmentation techniques. Next, we explain how we addressed these challenges. (1) Careful identification of appropriate data augmentation methods: Unlearnable data are carefully crafted, and it is essential to identify which data augmentation methods can most effectively attack them. However, due to the wide variety of existing data augmentation methods, the search space is vast, making an exhaustive search impractical. Therefore, we focused on nonlinear transformations, as they can attack both linear and nonlinear perturbations added to unlearnable data. To select the most effective among these nonlinear transformations, we propose a heuristic approach. (2) The potential requirement for multiple data augmentation methods to mitigate the effects of unlearnable perturbations: In our experiments, we observed that a single transformation may not be sufficient to effectively attack unlearnable data. To address this challenge, we opt to select a set of transformations rather than just one and expand the dataset by applying each transformation to it. (3) High computational cost due to the use of multiple data augmentation techniques: The learnable training set derived from unlearnable data is often five to six times larger than the original dataset, resulting in significantly increased training time. To manage computational costs, we limited data expansion to no more than six times the original dataset size. This constraint also influenced our selection of nonlinear transformations, as more complex transformations tended to increase execution time.

This paper addresses several challenges associated with attacking unlearnable data using nonlinear transformations. (1) Computational complexity: Nonlinear operations often involve more complex mathematical computations compared to linear ones and can be computationally intensive, especially for large images. To overcome this challenge, we developed an approach to leveraging  built-in Python functions to perform nonlinear transformations. (2) Generalization and overfitting: There is a risk of overfitting nonlinear transformations to specific datasets, and they may not generalize well across different image types. To address this generalization issue, our proposed algorithm searches for suitable transformations for each dataset individually. We identify nonlinear transformations tailored to each unlearnable dataset separately. Furthermore, instead of selecting a single transformation, we use multiple transformations per dataset to enhance robustness and generalization. (3) Interpretability: Outputs from nonlinear transformations are often less intuitive and harder to interpret. This is particularly evident in deep learning methods (e.g., CNNs), where it's challenging to relate transformations to specific features or image regions. To mitigate the difficulty in interpretability, we visually inspect a sample of the augmented data as shown in Fig. 3. This ensures that the applied transformations produce meaningful results. For example, transformations like Threshold Binary and Threshold Binary Inverse may not yield interpretable images for the CIFAR-10 dataset. (4) Parameter sensitivity: Nonlinear transformations often involve tunable parameters (e.g., thresholds), and their performance can vary significantly depending on these settings. To address this, key hyperparameters, such as the threshold values for Threshold Binary and Threshold Binary Inverse, and kernel sizes for operations like erode, dilate, and Gaussian blur—were selected through extensive experimentation. Each dataset required separate hyperparameter optimization to achieve optimal performance, guided primarily by test accuracy. (5) Difficult to analyze mathematically: While linear transformations are supported by well-established mathematical tools, nonlinear systems often lack general analytical solutions. This makes them harder to model or predict and complicates efforts to provide theoretical guarantees. As a result, we adopt a heuristic approach for optimizing nonlinear transformations. Using our selection method, we identify and apply nonlinear transformations such as erosion, dilation, and pixel-level manipulations—techniques that have been relatively unexplored in the context of unlearnable datasets.

\subsection{Nonlinear Transformations} \label{Appendix:DataAug}
\textbf{Threshold binary:} We know that a single number represents the pixel value for a gray image, whereas three numbers on the RGB scale represent the pixel value of a colored image. Although two different types of pixel values are used in gray and colored images, respectively, there is no difference between both types of images when the threshold binary approach is applied. First, we need to define a threshold value and the maximum value of a pixel. When a pixel value is lower than the predefined threshold, the pixel value will be zero. Otherwise, the pixel will be set to the maximum value. For grey images with a pixel value of $a$, a threshold value of $t$, and a maximum value of $m$, let $npv$ represent a new pixel value. Then, based on the threshold binary approach, $npv$ is defined as $0$ if $a \leq t$ or $m$ if $a>t$~\cite{harris2020understanding}.

Similarly, for colored images with a pixel value of $(r,g,b)$, a threshold value of $t$, and a maximum value of $m$, $npv= (nr,ng,nb)$ is the new pixel value, where $nr$, $ng$, and $nb$ are the new values of $r, g$, and $b$, respectively. Thus, according to the threshold binary approach, they are defined as: 

\begin{equation}
\begin{array}{cc}
    nr= \begin{cases}
        0, & \text{if } r \leq t \\
        m, & \text{if } r > t,
    \end{cases} &
    ng= \begin{cases}
        0, & \text{if } g \leq t \\
        m, & \text{if } g > t,
    \end{cases} \\
    \multicolumn{2}{c}{
        nb= \begin{cases}
            0, & \text{if } b \leq t \\
            m, & \text{if } b > t
        \end{cases}
    }
\end{array}
\end{equation}

We use threshold binary only for normalized datasets generated by NTGA. The pixel values for those images range from 0 to 1; therefore, we used 1 as our maximum value, corresponding to a value of 255 on non-normalized images~\cite{geeksforgeeks1}. The \texttt{cv2.threshold} function with the threshold method argument is set to \texttt{cv2.THRESH\_BINARY}, and it is used for the experiment. See Fig.~\ref{fig2} for an example image based on different data transformations. 

\textbf{Threshold binary inverse:} In OpenCV, the threshold binary inverse function has the same principle as the threshold binary function~\cite{harris2020fmix}. The only difference is that the pixel will receive a zero value when it is higher than the threshold value; otherwise, it will receive a maximum value. The cv2.threshold function with the threshold method argument is set to \texttt{cv2.THRESH\_BINARY\_INV}, and it is used for the experiment. The first three arguments are the same as for the threshold binary function. Please see Fig.~\ref{fig2} for an example image based on this data argument technique. 

\begin{figure*}[ht]
\centering
\includegraphics[scale=0.5]{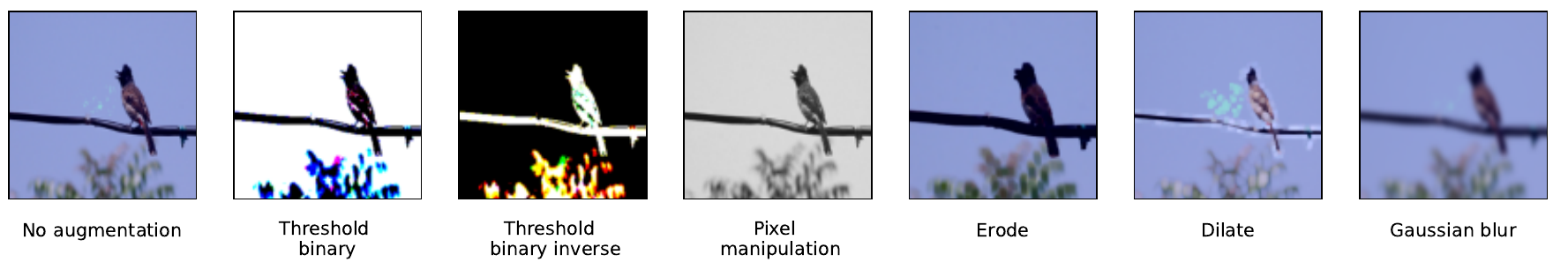}
\caption{An illustration of nonlinear transformation techniques.}
\label{fig2}
\end{figure*}

\begin{figure}
\vspace{-0.8in}
\includegraphics[scale=0.6,trim = {0.7cm 1cm 1cm 1cm}]{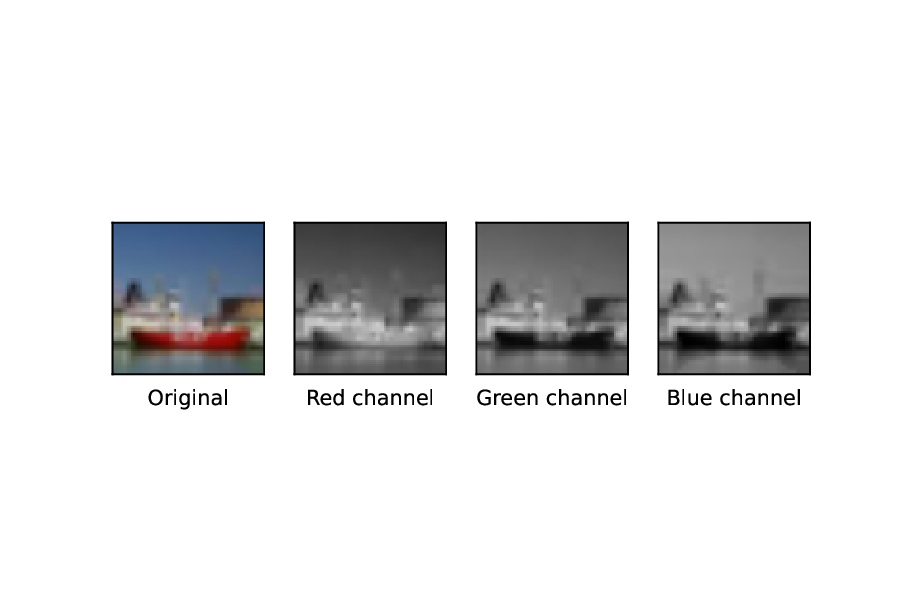}
\vspace{-0.5in}
\caption{An illustration of the color channels.}
\label{figchannel}
\end{figure}

\textbf{Color channel manipulation:} It is another technique typically used for pixel manipulation in our experiment. An image consists of multiple pixels that contain information about the color of a minute area in that image. Each pixel of a colored image is composed of three values representing the intensity of blue (b), green (g), and red (r) light colors, respectively. However, each pixel of a grey image has only one value representing the light intensity of an image. Color channel manipulation is about changing the color value of one or more color channels. In our experiments, we used color channels to manipulate the pixels based on the code available for operations on images~\cite{ColorChannelManipulation}. For instance, when using the cv2.merge((b,b,b)), all three channels' values are replaced by the blue channel's value b. In the original image shown in Fig. \ref{figchannel}, the sky is blue, indicating that the blue channel value is the largest among all three values for each pixel in the sky area. When cv2.merge((b,b,b)) replaces three channels with that value $b$, the blue sky in the original image looks paler in the last image (as shown in Fig. \ref{figchannel}) because the purer blue it is, the closer the value is to (255, 255, 255), indicating a white pixel.

Further, sky pixels have low red and green channel values. When those values are used for all three channels shown in the second and third images of Fig. \ref{figchannel}, respectively, pixels will get closer to black color since the black is indicated by (0, 0, 0). Color channel manipulations efficiently generate additional images with a negligible computation overhead effort.

\textbf{Erode:} This is a commonly used image-processing technique introduced in mathematical morphology. This technique was initially defined for binary images (black and white), but it was later extended to grayscale images. Erosion reduces bright areas of an image and replaces them with dark regions \cite{DBLP:journals/pami/HaralickSZ87}. Consider $A$ and $B$ as sets in $\mathbb{Z}^2$, where $A$ is considered as the coordinates of an input image, and B denotes the structuring element or the shape parameter. \cite{DBLP:journals/pami/HaralickSZ87} denote the translation of $B$ by $x \in \mathbb{Z}^2$ as $(B)_x$ as follows:
\begin{equation}
\label{eq:translation}
(B)_x =\{c \in Z^2| c=b+x , \exists b \in B \}.
\end{equation}
Then, \cite{DBLP:journals/pami/HaralickSZ87} defined erosion in the following way.
\begin{definition}
\label{def1}
The erosion of A by B  $(A \ominus B)$ is defined as:
\begin{equation}
A \ominus B =\{x| (B)_x \subseteq A  \}.
\end{equation}
\end{definition}

The erosion of $A$ by $B$ is the set of points $x$ such that the translation of $B$ by $x$ is contained in $A$. In other words, the erosion of $A$ by $B$ includes the points that translate B in a way such that translated B does not have any points outside A \cite{gonzalez2008digital}. In a programming setting, structuring element B is known as a kernel. Based on the kernel, we can control the severity of erosion. We used \texttt{cv2.erode} in OpenCV for the erode transformation, which has three main arguments. The first argument is a base image ($A$). The second argument is a kernel, $B$. We define the kernel as an all-ones matrix that slides across $A$. If all pixels under the kernel are 1, the original pixel value will be converted to 1; otherwise, the pixel value is 0. In that way, the white region of an original image will be reduced. The third argument in this function is the number of iterations. It specifies how many times we want to perform this transformation.

\textbf{Dilate:} Dilate is the dual transform of erode. Dilation is also mainly introduced to binary images. As the name reveals, a dilate transformation grows or expands the bright region of an image into a black area in the background of an image \cite{DBLP:journals/pami/HaralickSZ87}. \cite{DBLP:journals/pami/HaralickSZ87}, define dilation transformation as follows.
\begin{definition}
\label{def2}
The dilation of $A$ by $B$ $(A \oplus B)$ is defined by:
\begin{equation}
A \oplus B =\{c \in Z^2| c=a+b, \exists a \in A \mbox{ and } \exists b \in B \}.
\end{equation}
\end{definition}

In our experiments, we performed a dilate transformation based on~\texttt{cv2.dilate} in OpenCV. A dilate function also has the same arguments as an erode function, i.e., a base image, a kernel, and the number of iterations. Like the erode function, the kernel is specified as an all-ones matrix and slides cross $A$. However, it does not perform the same way as erode. If at least one of the pixels under the kernel is 1, the original pixel value will be converted to 1; otherwise, the pixel value is 0. This transformation will expand the white region of an image.

\textbf{Gaussian blur:} The Gaussian blur transformation of a pixel value is computed by taking the weighted average of neighboring pixel values. A Gaussian blur transformation is carried out by the convolution that involves a kernel generated using a Gaussian function. A Gaussian function is defined as follows:
\begin{equation}\label{gaussian}
h(x,y)=K \exp \Big\{\frac{-(x^2+y^2)}{2\sigma^2} \Big\},
\end{equation}
where $K$ is a normalization constant, $x$ and $y$ are distances from the original pixel to its neighbor pixel in horizontal and vertical axes, respectively, and $\sigma$ denotes the standard deviation of ($x$, $y$), which controls the intensity of the blur. For our experiments, we conducted a Gaussian blur transformation by using \texttt{cv2.GaussianBlur} in OpenCV. Similar to dilate and erode, we need to specify the size of a kernel before applying a Gaussian blur transformation. For example, based on extensive experiments, we chose a kernel with length of 55 and width of 5 in this research.

\section{Experimental Evaluation} \label{sec:evaluation}

We have empirically demonstrated, through extensive experiments and analysis, that major data protection approaches-namely, Deepconfuse \cite{DBLP:conf/nips/FengCZ19}, error-minimizing noise approach \cite{DBLP:conf/iclr/HuangME0021}, error-maximizing noise approach \cite{DBLP:journals/corr/abs-2106-10807}, NTGA \cite{yuan2021neural}, synthetic approach~\cite{yu2021indiscriminate}, autoregressive noise approach \cite{DBLP:journals/corr/abs-2206-03693}, OPS~\cite{DBLP:conf/iclr/WuCXH23}, SEP~\cite{chen2022self}, EntF~\cite{DBLP:conf/iclr/00020000023}, REM~\cite{DBLP:conf/iclr/FuHLST22}, Hypocritical~\cite{DBLP:conf/nips/TaoFYHC21}, TensorClog~\cite{DBLP:journals/access/ShenZM19}, and PUE~\cite{wang2024provably} —can be effectively circumvented by leveraging the proposed framework outlined in Section~\ref{sec:methodology}. We have extensively investigated the learnability of the 13 data protection approaches using the CIFAR-10~\cite{krizhevsky2009learning} dataset.  We selected CIFAR-10 since it is the standard dataset used by many unlearnable example researchers. Each unlearnable training dataset consists of 50000 images, except for the NTGA dataset released by~\cite{yuan2021neural}, which contains 40000 training images. The test dataset includes only 10000 clean images loaded from TensorFlow Datasets. 
Additionally, we apply our framework to two other benchmark datasets: MNIST~\cite{lecun-mnisthandwrittendigit-2010} and ImageNet \cite{DBLP:conf/cvpr/DengDSLL009} generated by NTGA. Those results are presented in Section \ref{discussion}. 

In summary, our experimental details are organized as follows:
\begin{itemize}
    \item \textbf{Datasets used:} We  specify the use of CIFAR-10, MNIST, and ImageNet datasets in Section~\ref{deepconfuse} and~\ref{discussion}.
    \item \textbf{Models employed:} Detailed model architectures are provided in Tables~\ref{t:NTGAmodels} and~\ref{t:Othermodels}. These include: VGG16, VGG19, and ResNet152 pretrained models with specific architectural modifications such as added fully connected and dropout layers. Each dataset and attack scenario includes a clearly defined model setup (see Table~\ref{t:6} for a summary, and individual subsections in Section~\ref{sec:evaluation} for full descriptions).

    \item \textbf{Baseline methods for comparison:} we compare our method against: The Orthogonal Projection Attack (OPA) from~\cite{DBLP:journals/corr/abs-2305-19254} (a linear transformation-based baseline). Adversarial training as an additional baseline defense strategy (Table~\ref{t:5}, Column 7). Table~\ref{t:5} presents comprehensive comparisons showing accuracy improvements across  13 data protection approaches using these baselines. Additionally, we have compared our approach with four other attack methods in Section~\ref{sec.compare}.
\end{itemize}

\subsection{The Proposed Framework on the 13 Unlearnable Approaches}
\label{deepconfuse}

We  aim to reveal that the unlearnable CIFAR-10 datasets created by the 13 popular data protection approaches, i.e., NTGA~\cite{yuan2021neural}, Deepconfuse~\cite{DBLP:conf/nips/FengCZ19}, error-minimizing~\cite{DBLP:conf/iclr/HuangME0021}, error-maximizing~\cite{DBLP:journals/corr/abs-2106-10807}, synthetic~\cite{yu2021indiscriminate}, autoregressive~\cite{DBLP:journals/corr/abs-2206-03693}, OPS~\cite{DBLP:conf/iclr/WuCXH23}, SEP~\cite{chen2022self}, EntF~\cite{DBLP:conf/iclr/00020000023}, REM~\cite{DBLP:conf/iclr/FuHLST22}, Hypocritical~\cite{DBLP:conf/nips/TaoFYHC21}, TensorClog~\cite{DBLP:journals/access/ShenZM19}, and PUE~\cite{wang2024provably} are also vulnerable to our proposed approach. In our work, the datasets generated by Deepconfuse, error-minimizing, error-maximizing, and synthetic approaches were obtained from \cite{yu2021indiscriminate}. \cite{yuan2021neural} publicly released three unlearnable datasets: MNIST, CIFAR-10, and ImageNet on Kaggle. We utilized these data to demonstrate the vulnerability of NTGA perturbations to our approach. The datasets created by other approaches- autoregressive, OPS, SEP, EntF, REM, Hypocritical, TensorClog, and PUE were generated using the available code in their respective GitHub repositories.

\begin{table*}[ht]
\centering
\caption{The baseline test accuracy is given in Column 2 whose results are extracted from those papers that presented their respective approaches given Column 1.
Our approach in column 4 performs higher test accuracy than OPA, a linear transformation technique, given in Column 3 for all the 13 approaches except OPS and PUE, where six of them are remarkably better. Our experimental results in Column 4 also suggest that these 13 data protection approaches remain vulnerable to nonlinear transformations. While Column 5 gives the performance difference between our nonlinear transformation approach and OPA (a linear transformation approach) in~\cite{DBLP:journals/corr/abs-2305-19254}, Column 6 presents the performance improvement percentages of our approach compared to OPA. Our research also adopted the adversarial training method  in~\cite{DBLP:conf/iclr/FuHLST22} for model training, and Column 7 gives resulting test accuracy for the trained models. Columns 6 and 7 clearly demonstrates that our approach can achieve comparable test accuracy like the adversarial training method, while our approach performs much better for OPS.}
 \label{t:5}
\renewcommand{\arraystretch}{0.5}
\begin{tabular}{|>{\raggedright\color{black}}P{2.5cm}|>{\color{black}}P{1.6cm}|>{\color{black}}P{1.5cm}|>{\color{black}}P{1.6cm}|>{\color{black}}P{1.8cm}|>{\color{black}}P{1.8cm}|>{\arraybackslash\color{black}}P{1.6cm}| }
 \hline
 Data protection approach & 
No defense & OPA & Our approach& Performance difference & \% Improvement &Adversarial training\\
 \hline
 NTGA & 40\% & 61.54\% & 87.38\%  & 25.84\% &41.98\% &   84.75\%\\
 \hline
 Deepconfuse & 29\% & 75.28\% & 86.93\%  & 11.65\% & 15.48\% &  84.92\%\\
 \hline
 Error-minimizing &20\% & 69.18\%& 85.26\%  & 16.08\% & 23.24\% &   79.68\% \\
 \hline
 Error-maximizing &6\% & 75.57\% & 90.75\%  &  15.18\% & 20.09\% &   85.36\%\\
 \hline
 Synthetic &  13\% & 87.90\% & 88.49\% &  0.59\% &0.67\% &  86.74\%\\
 \hline
 Autoregressive & 11\% &25.59\% & 85.66\% & 60.07\% &234.74\% &  80.54\% \\
\hline
OPS&    15\% & 88.10\% &   87.76\%  &-0.34\% & -0.39\% & 13.01\%\\
 \hline
     SEP&     23\%& 87.28\% &     88.27\%&   0.99\%& 1.13\% &  87.20\%\\
 \hline
     EntF&     71\%&    85.67\% & 87.27\%& 1.6\% &  1.83\%   & 82.30\% \\
 \hline
     REM&     27\% &  34.78\%&    86.21\%& 51.43\% & 147.87\% & 49.51\%\\
 \hline
     Hypocritical&    19\% & 89.1\%&      89.68\%&  0.58\% & 0.65\% & 80.41\%\\
 \hline
  TensorClog& 48\% &  88.05\%&    90.01\%&   1.96\%& 2.22\% & 85.16\%\\
 \hline
  PUE & 11\% &  90.09\% &  85.89\% &  -4.2\%& -4.66\% & 75.35\% \\
 \hline
 \end{tabular} 

\end{table*}

For all the datasets, we used Tensorflow's pretrained VGG models initialized with ImageNet weights. Table~\ref{t:6} explicitly mentions the nonlinear transformations and model specifications used in each unlearnable dataset based on the proposed framework, where we conducted extensive experiments for all those 13 data protection approached studied in this paper. Column 2 gives the transformations used to expand the training dataset. In this table, Column 3 includes the attributes used for Keras ImageDataGenerator to conduct more transformations during the training process. Last three columns specify the model specifications.

Table~\ref{t:5} summarizes our study's main findings, with column 2 displaying test accuracies mostly below 30\% for models trained on these data protection approaches. To demonstrate the ability to learn from the so-called unlearnable dataset, we employed our nonlinear transformation approach detailed in Section~\ref{sec:methodology}. Its resulting accuracy is given in Column 4. Column 3 shows the test accuracy of the models trained on the same unlearnable datasets using linear transformation technique, specifically OPA in~\cite{DBLP:journals/corr/abs-2305-19254}. We employed the code provided in their GitHub repositories with default settings, such as a non-pretrained ResNet18 model under PyTorch platform. Column 5 gives the difference between the test accuracies of our approach and the OPA. The performance difference is shown as a percentage in Column 6. The last column shows the accuracy after PGD adversarial training \cite{DBLP:conf/iclr/FuHLST22}, which is another widely used method to break unlearnabiilty. Our approach in column 4 performs higher test accuracy than OPA, a linear transformation technique, given in Column 3 for all the 13 approaches except OPS and PUE, where six of them are remarkably better. Our experimental results in Column 4 also suggest that these 13 data protection approaches remain vulnerable to nonlinear transformations.

\begin{table*}[ht]
\small
\caption{The model architecture used for training each unlearnable dataset crafted by NTGA.}
\label{t:NTGAmodels}
\scriptsize
\centering
\begin{tabular}{c c}
    \begin{minipage}{0.48\textwidth}
        \begin{subtable}{\textwidth}
        \caption{MNIST}
            \label{fig:modelmnist}
            \begin{tabular}{|c c c|} 
                \hline
                Layer & Output shape & Activation function \\  
                \hline
                Input  & 28 x 28 x 1 &  \\ 
                \hline
                Convolutional   & 28 x 28 x 32 & ReLU \\
                \hline
                Max pool & 14 x 14 x 32  &  \\
                \hline
                Convolutional & 14 x 14 x 64 & ReLU  \\
                \hline
                Max pool & 7 x 7 x 64 &  \\ 
                \hline
                 &  Flatten & \\
                \hline
                FC & 1024 & ReLU\\
                \hline
                 & Dropout (0.25) & \\
                \hline
                FC & 10 & Softmax\\
                \hline
            \end{tabular}
        \end{subtable}

        \begin{subtable}{\textwidth}
        \caption{CIFAR-10}
            \label{fig:modelcifar}
            \begin{tabular}{|c c c|} 
                \hline
                Layer & Output shape & Activation function \\  
                \hline
                Input  & 32 x 32 x 3 &   \\ 
                \hline
                2 x Convolutional   & 32 x 32 x 64 & ReLU \\
                \hline
                Max pool & 16 x 16 x 64  &  \\
                \hline
                2 x Convolutional & 16 x 16 x 128 & ReLu  \\
                \hline
                Max pool & 8 x 8 x 128 &  \\ 
                \hline
                4 x Convolutional & 8 x 8 x 256 & ReLU \\
                \hline
                Max pool & 4 x 4 x 256 & \\ 
                \hline
                4 x Convolutional & 4 x 4 x 512 & ReLU\\
                \hline
                Max pool & 2 x 2 x 512 &\\
                \hline
                4 x Convolutional & 2 x 2 x 512 & ReLU\\
                \hline
                Max pool & 1 x 1 x 512 & \\
                \hline
                & Flatten &\\
                \hline
                FC & 1024 & ReLU \\
                \hline
                & Dropout (0.3)  & \\
                \hline
                FC & 512 & ReLU \\
                \hline
                & Dropout (0.3) & \\
                \hline
                FC & 64 & ReLU\\
                \hline 
                FC & 10 & Softmax\\
                \hline
            \end{tabular}
        \end{subtable}
    \end{minipage}
    &
    \begin{minipage}{0.48\textwidth}
        \begin{subtable}{\textwidth}
        \caption{ImageNet}
     \label{fig:ImageNet model}
        \begin{tabular}{| c c c| }
     \hline
     Layer  & Output shape & Activation function \\
     \hline
     Input & 224 x 224 x 3 & \\
     \hline
     7 x 7, 64, / 2 & 112 x 112 x 64 & ReLU\\  
     \hline
     3 x 3 max pool, / 2 & 56 x 56 x 64 & \\ 
     \hline $\begin{bmatrix} 1 \times 1 ,64\\ 3 \times 3 ,64\\ 1 \times 1 ,256 \end{bmatrix}$ x 3 & 56 x 56 x 64 & ReLU\\[0.5cm]
     \hline
     $\begin{bmatrix} 1 \times 1 ,128\\ 3 \times 3 ,128\\ 1 \times 1 ,512
     \end{bmatrix}$ x 8 & 28 x 28 x 128 & ReLU\\[0.5cm]
     \hline
     $\begin{bmatrix} 1 \times 1 ,256\\ 3 \times 3 ,256\\ 1 \times 1 ,1024
     \end{bmatrix}$ x 36 & 14 x 14 x 256 & ReLU \\[0.5cm]
     \hline
     $\begin{bmatrix} 1 \times 1 ,512\\ 3 \times 3 ,512\\ 1 \times 1 ,2048
     \end{bmatrix}$ x 3 & 7 x 7 x 512 & ReLU \\[0.5cm]
     \hline
     $\begin{bmatrix} 1 \times 1 ,512\\ 3 \times 3 ,512\\ 1 \times 1 ,2048
     \end{bmatrix} $ x 3 & 1 x 1 x 512 & ReLU  \\[0.5cm] \hline
     Average pool & 1 x 1 x 512 & Softmax\\ \hline
     & Flatten &\\ \hline
     FC & 1024 & ReLU \\ \hline
     & Dropout (0.3) &\\ \hline
     FC & 512 & ReLU\\ \hline
     FC & 256 & ReLU \\ \hline
     FC & 128 & ReLU \\ \hline
     FC & 64 & ReLU \\ \hline
     FC & 2 & Softmax \\ \hline
     \end{tabular}
    
        \end{subtable}
    \end{minipage} 
\end{tabular}

\end{table*}

\begin{table}[ht]
\small
\scriptsize
\caption{The model architectures used for training the unlearnable CIFAR-10 images crafted by each approach.}
   \label{t:Othermodels}
    \begin{subtable}[b]{0.49\textwidth}
        \centering
         \caption{Deepconfuse}
    \label{fig:dcarchitectur}
        \begin{tabular}{|c c c|} 
    \hline
    Layer & Output shape & Activation function \\  
    \hline
    \multicolumn{3}{|c|}{VGG19} \\
    \hline
    \multicolumn{3}{|c|}{Flatten}\\
    \hline
    FC & 2048 & ReLU \\
    \hline
    \multicolumn{3}{|c|}{Dropout (0.3)} \\
    \hline
    FC & 1024 & ReLU \\
    \hline
    \multicolumn{3}{|c|}{Dropout (0.3)} \\
    \hline
    FC & 512 & ReLU\\
    \hline
    FC & 256 & ReLU\\
    \hline
    FC & 128 & ReLU\\
    \hline
    FC & 64 & ReLU\\
    \hline 
    FC & 10 & Softmax\\
    \hline
    \end{tabular}
    \end{subtable}
    \begin{subtable}[b]{0.49\textwidth}
        \centering
        \caption{Error-minimizing}
    \label{fig:eminarchitecture}
       \begin{tabular}{|c c c|} 
    \hline
    Layer & Output shape & Activation function \\  
     \hline
    Input  & 32 x 32 x 3 &   \\ 
    \hline
    2 x Convolutional   & 32 x 32 x 64 & ReLU \\
    \hline
    Max pool & 16 x 16 x 64  &  \\
    \hline
    2 x Convolutional & 16 x 16 x 128 & ReLu  \\
    \hline
    Max pool & 8 x 8 x 128 &  \\ 
    \hline
    3 x Convolutional & 8 x 8 x 256 & ReLU \\
    \hline
    Max pool & 4 x 4 x 256 & \\ 
    \hline
    3 x Convolutional & 4 x 4 x 512 & ReLU\\
    \hline
    Max pool & 2 x 2 x 512 &\\
    \hline
    3 x Convolutional & 2 x 2 x 512 & ReLU\\
    \hline
    Max pool & 1 x 1 x 512 & \\
    \hline
    & Flatten &\\
    \hline
    FC & 256 & ReLU \\
    \hline
    FC & 10 & Softmax\\
    \hline
  \end{tabular}
    
    \end{subtable}
%
   \begin{subtable}[b]{0.31\textwidth}
       \centering
       \caption{Error-maximizing and REM }
    \label{fig:emaxarchitectur}
        \begin{tabular}{|c c c|} 
    \hline
    Layer & Output shape & Activation function \\  
       \hline
    \multicolumn{3}{|c|}{VGG16} \\
    \hline
    \multicolumn{3}{|c|}{Flatten} \\
    \hline
    FC & 2048 & ReLU \\
    \hline
    FC & 1024 & ReLU \\
    \hline
    FC & 512 & ReLU \\
    \hline
    FC & 256 & ReLU\\
    \hline
    FC & 180 & ReLU\\
    \hline
    FC & 128 & ReLU \\
    \hline
    FC & 64 & ReLU\\
    \hline 
    FC & 10 & Softmax\\
    \hline
    \end{tabular}
    
   \end{subtable}
   
   \begin{subtable}[b]{0.31\textwidth}
       \centering
       \caption{Synthetic, autoregressive, OPS, EnF, Hypocritical, and TensorClog}
    \label{fig:synarchitecture}
        \begin{tabular}{|c c c|} 
    \hline
    Layer & Output shape & Activation function \\  
    \hline
    \multicolumn{3}{|c|}{VGG19} \\
    \hline
    \multicolumn{3}{|c|}{Flatten}\\
    \hline
    FC & 1024 & ReLU \\
    \hline
    FC & 512 & ReLU\\
    \hline
    FC & 128 & ReLU\\
    \hline
    FC & 64 & ReLU\\
    \hline 
    FC & 10 & Softmax\\
    \hline
    \end{tabular}
    
   \end{subtable}
   \begin{subtable}[b]{0.31\textwidth}
     \caption{SEP and PUE}
    \label{fig:separchitecture}
       \centering
        \begin{tabular}{|c c c|} 
    \hline
    Layer & Output shape & Activation function \\  
    \hline
    \multicolumn{3}{|c|}{VGG16} \\
    \hline
    \multicolumn{3}{|c|}{Flatten}\\
    \hline 
    FC & 10 & Softmax\\
    \hline
    \end{tabular}
   \end{subtable}
   
\end{table}

\begin{table*}[h]
\footnotesize
\caption{Summary of transformations and model specifications used to make these unlearnable CIFAR-10 images learnable.}
 \label{t:6}
\begin{tabular}
{|>{\raggedright}P{1.5cm} |>{\raggedright}P{3cm}|>{\raggedright}P{5cm}|>{\raggedright}P{1.3cm}|>{\raggedright}P{1.1cm}|>{\raggedright\arraybackslash}P{0.8cm}|}
 \hline
 Protection approach & Nonlinear Transformations  &  Attributes for Keras ImageDataGenerator & Model Architecture & Learning rate & \# of epochs \\
 \hline
 NTGA & color channel manipulation (thrice) & rotation range of 7, zoom range of 0.3, width and height shift range of 0.35, horizontal flip, shear range of 0.4 &  Table~\ref{fig:modelcifar} & 0.008 & 40 \\
 \hline
 Deepconfuse & dilate, erode, and color channel manipulation (twice) & rotation range of 7, width shift range of 0.3, horizontal flip, zoom range of 0.1 & Table~\ref{fig:dcarchitectur} & 0.007 & 80 \\
 \hline
 Error-minimizing & dilate, erode, color channel manipulation (twice), and Gaussian blur &  rotation range of 7, height and width shift range of 0.3, horizontal flip,  zoom range of 0.1 & Table~\ref{fig:eminarchitecture}  & 0.006 & 80 \\
 \hline
 Error-maximizing & dilate, erode, and color channel manipulation (twice) & rotation range of 10,  shear range of 0.4, height and width shift range of 0.4, horizontal flip, zoom range of 0.4 & Table~\ref{fig:emaxarchitectur}  & 0.008 & 80  \\
 \hline
 Synthetic & dilate, erode, and color channel manipulation (twice) &rotation range of 7, width and height shift range of 0.3, shear range of 0.4, horizontal flip, zoom range of 0.4 & Table~\ref{fig:synarchitecture}  & 0.007 & 40 \\
 \hline
 Autoregressive & dilate, erode, and color channel manipulation (twice) & rotation range of 7, horizontal flip, zoom range of 0.1 & Table~\ref{fig:synarchitecture}  & 0.001 & 10
 \\
 \hline
  OPS &  dilate, erode, and color channel manipulation (twice) &  rotation range of 7, width and height shift range of 0.3, horizontal flip, zoom range of 0.1 &  Table~\ref{fig:synarchitecture}  &  0.001  &  30 \\
 \hline
  SEP &  color channel manipulation (twice) &  rotation range of 7, width and height shift range of 0.1, horizontal flip, zoom range of 0.1 &  Table~\ref{fig:separchitecture}  &  0.001  &  50 \\
 \hline
  EntF &  erode and color channel manipulation &  rotation range of 7, width and height shift range of 0.3, horizontal flip, zoom range of 0.1 &  Table~\ref{fig:synarchitecture}  &  0.001  &  30 \\
 \hline
   REM &  erode and color channel manipulation (twice) &  rotation range of 7, width shift range of 0.15, and height shift range of 0.2, horizontal flip, zoom range of 0.1 & Table~\ref{fig:emaxarchitectur} &  0.006  &  80 \\
 \hline
   Hypocritical &  dilate, erode and color channel manipulation &  rotation range of 7, width and height shift range of 0.3, horizontal flip, zoom range of 0.1 &  Table~\ref{fig:synarchitecture}  &  0.001  &  30 \\
 \hline
  TensorClog &  dilate, erode and color channel manipulation &  rotation range of 7, width and height shift range of 0.3, horizontal flip, zoom range of 0.1 &  Table~\ref{fig:synarchitecture}  &  0.001  &  30 \\
  \hline
  PUE & dilate, erode and color channel manipulation (thrice) & rotation range of 7, width and height shift range of 0.1, horizontal flip, zoom range of 0.1 &  Table~\ref{fig:separchitecture} & 0.001  & 50 \\
 \hline
 \end{tabular} 
\end{table*}

Table~\ref{t:6} presents the experimental settings used to break each unlearnable dataset with our approach. Column 2 gives the nonlinear transformation applied on the unlearnable dataset. Column 2 shows the nonlinear transformation applied to each unlearnable dataset. The training dataset was expanded depending on the number of transformations applied. For example, the NTGA dataset was expanded threefold by applying a pixel manipulation transformation three times. Additionally, we utilized Keras’s built-in ImageDataGenerator on the training dataset for further augmentation. The specific attributes used for ImageDataGenerator are listed in Column 3 of Table \ref{t:6}. The model architecture used for training is provided in Column 4. For instance, refer to Table~\ref{fig:modelcifar} for a detailed view of the model architecture used on NTGA dataset. As a baseline model, we conducted experiments based on a Visual Geometry Group (VGG) in~\cite{DBLP:journals/corr/SimonyanZ14a}. There are several variants in VGG models depending on the number of convolutional layers. After thorough experiments with VGG16, VGG19, and ResNet50, we selected the VGG19 model with ImageNet pretrained weights from Keras for its superior baseline performance. Hyperparameters such as learning rate and number of training epochs were carefully determined through comprehensive experimentation. These values are provided in Columns 5 and 6 of Table~\ref{t:6}.

\textbf{NTGA~\cite{yuan2021neural}:} We illustrate the effectiveness of our proposed framework in countering the unlearnable CIFAR-10 data generated by NTGA~\cite{yuan2021neural}. As reported in~\cite{yuan2021neural}, the lowest test accuracy for the unlearnable CIFAR-10 data is approximately 41\%.
We aim to demonstrate that the unlearnable CIFAR-10 crafted by NTGA becomes learnable applying our proposed framework. We used the proposed nonlinear transformation based framework to increase our training dataset size, enhancing the model's resilience to image transformations. The model resulted in a test accuracy of 87.38\%,  and a training accuracy of 92.95\%. Training the same model without nonlinear transformations yielded a test accuracy of 39.24\%, underscoring the substantial 48\% accuracy improvement achieved through nonlinear transformations. Using the linear transformation technique in~\cite{DBLP:journals/corr/abs-2305-19254} (OPA), we obtained a model with a test accuracy of 61.54\%.

\textbf{Deepconfuse~\cite{DBLP:conf/nips/FengCZ19}:} 
As reported in \cite{DBLP:conf/nips/FengCZ19}, models trained on the CIFAR-10 unlearnable dataset generated by Deepconfuse exhibit an average test accuracy of 29\%. 
The model achieved by applying our approach resulted in a test accuracy of 86.93\%. Training the same model without nonlinear transformations yielded a test accuracy of only 30.68\%, consistent with the findings in \cite{DBLP:conf/nips/FengCZ19}. Using OPA on the same dataset achieved a ResNet18 model with test accuracy of 75.28\%. Hence, our nonlinear transformation approach yields a model that has a 15.48\% improvement compared to the model trained with OPA. This result reaffirms that our approach can render unlearnable data learnable, showcasing the efficacy of nonlinear transformations.

\textbf{Error-minimizing~\cite{DBLP:conf/iclr/HuangME0021}:} 
To demonstrate the learnability of the unlearnable CIFAR-10 dataset generated by the error-minimizing approach, we augmented the dataset size to five times its original size using transformations detailed in Table \ref{t:6}. The model resulted in a test accuracy of 85.26\%. However, utilizing the same model architecture without nonlinear transformation techniques led to a test accuracy of only 28.62\%, akin to the result reported in \cite{DBLP:conf/iclr/HuangME0021}. Using OPA code on the same dataset, we achieved a model with test accuracy of 69.18\%. Therefore, the model resulted using our approach showed 16.08\% improvement in the test accuracy.

\textbf{Error-maximizing~\cite{DBLP:journals/corr/abs-2106-10807}:}  According to the study \cite{DBLP:journals/corr/abs-2106-10807}, the lowest test accuracy of the model trained on the unlearnable CIFAR-10 data created by the error-maximizing approach is 6.25\%. Our goal is to demonstrate that this dataset is still learnable by achieving a test accuracy of 85\%. The model obtained using our approach resulted in a test accuracy of 90.75\%. The linear separable technique was only able to obtain a model with test accuracy of 75.57\%. 

\textbf{Synthetic~\cite{yu2021indiscriminate}:} 
The model trained on unlearnable CIFAR-10 examples crafted by the synthetic approach achieved a test accuracy of 13.54\% reported in~\cite{yu2021indiscriminate}. However, we demonstrate that these seemingly unlearnable CIFAR-10 examples from the synthetic approach are indeed learnable, achieving an 88\% test accuracy with appropriate nonlinear transformations. Conversely, training the same model without nonlinear transformations led to a 42.7\% test accuracy. Since the synthetic approach involves class-wise perturbations, the linear separability technique is reported to be effective in the breaking synthetic approach~\cite{DBLP:journals/corr/abs-2305-19254}. This fact is confirmed by the test accuracy of 87.9\% we obtained after employing OPA.

\textbf{Autoregressive~\cite{DBLP:journals/corr/abs-2206-03693}:} 
As per~\cite{DBLP:journals/corr/abs-2305-19254}, autoregressive perturbations are not linearly separable, making them difficult to break using OPA. In our experiments, we utilized an unlearnable dataset with autoregressive perturbation of $\epsilon=1$, reported to have a test accuracy of 11.75\%. After applying our approach on the autoregressive dataset, we achieved a model with a test accuracy of 86.66\%. As expected, OPA was able to achieve a model with a low test accuracy of 25.59\%.

\textbf{OPS~\cite{DBLP:conf/iclr/WuCXH23}:} 
As reported in~\cite{DBLP:conf/iclr/WuCXH23}, the ResNet18 model trained on unlearnable examples generated by OPS perturbations achieved a test accuracy of 15.56\%. Similar to synthetic perturbations, OPS is also a type of class-wise perturbations that is highly vulnerable to the linear transformation technique performed by OPA in~\cite{DBLP:journals/corr/abs-2305-19254}. After applying our approach on the OPS dataset, we achieved a model with a test accuracy of 87.76\%. The ResNet18 model trained using OPA achieved a slightly better test accuracy of 88.10\%, indicating that linear transformations are more appropriate for breaking OPS perturbations.

\textbf{SEP~\cite{chen2022self}:} 
We tested the SEP data protection approach using our framework. Employing their GitHub code, we generated the best-protected dataset, SEP-FA-VR, with a perturbation radius of 2/255. the VGG16 model (from the PyTorch code in their GitHub repository) trained on this dataset achieved a test accuracy of 24.88\%, consistent with results in~\cite{chen2022self}. However, using TensorFlow's pretrained VGG16 model with ImageNet weights on the same unlearnable dataset, without any transformations, we achieved a higher test accuracy of 83.81\%. This highlights the enhancing effect of pretrained models on learning from SEP-protected data. Further applying our nonlinear transformations approach boosted the test accuracy to 88.27\%. It is worth noticing that OPA also achieved a model with almost the same test accuracy of 87.28\%.

\textbf{EntF~\cite{DBLP:conf/iclr/00020000023}:} EntF is a recently proposed data protection approach created using entangled features. We generated unleanable CIFAR-10 dataset perturbed with EntF by employing their code on GitHub with default settings (perturbation radius of 8/255). As per~\cite{DBLP:conf/iclr/00020000023}, an adversarially trained model on an unlearnable dataset achieved a test accuracy of 71.57\%. Our aim, using the outlined approach, was to reach a model with an 85\% accuracy. Using our approach, we achieved a model with a test accuracy of 87.27\%. After applying OPA on the same dataset, we obtained a model with test accuracy of 85.67\%. Hence, our approach shows 1.83\% improvement compared to OPA.

\textbf{REM~\cite{DBLP:conf/iclr/FuHLST22}:} 
The CIFAR-10 dataset with REM noise is generated using their GitHub code with default settings (perturbation radius of 4/255). As reported in~\cite{DBLP:conf/iclr/FuHLST22}, the model trained on this dataset achieved a test accuracy of 27.09\%. After applying our framework, we achieved a model with a test accuracy of 86.21\%. This demonstrates the model's ability to learn from the ostensibly unlearnable data with REM noise. We then applied OPA on the dataset but obtained a lower test accuracy of 34.78\%, meaning that REM perturbations are not vulnerable to linear transformations. This fact is confirmed by the similar results in~\cite{DBLP:journals/corr/abs-2305-19254}. Hence, our nonlinear approach obtained a model with 51.43\% more accuracy than the model obtained using the linear approach. However, the model exhibits a training accuracy of 99.45\%. 

\textbf{Hypocritical~\cite{DBLP:conf/nips/TaoFYHC21}:}  Hypocritical perturbations is one of the data protection approaches discussed in~\cite{DBLP:conf/nips/TaoFYHC21}. We generated the class-wise Hypocritical perturbations since it is more effective than sample-wise perturbations. After applying the standard training method provided in their GitHub repository on the generated unlearnable CIFAR-10 dataset, the resulting model achieved a test accuracy of 18.59\%, consistent with the result reported in~\cite{DBLP:conf/nips/TaoFYHC21}. However, using the pretrained model in Table~\ref{fig:synarchitecture} without any transformations on the same dataset, we obtained a test accuracy of 79.77\%. After applying nonlinear transformations, we achieved a test accuracy of 89.12\%. The training accuracy is 93.90\%. After applying OPA, we obtained a model with a test accuracy of 86.79\%. It is reasonable that this dataset can be broken by the orthogonal projection method since it has class-wise perturbations.

\textbf{TensorClog~\cite{DBLP:journals/access/ShenZM19}:} According to~\cite{DBLP:journals/access/ShenZM19}, TensorClog perturbations can reduce the test accuracy of a model trained on CIFAR-10 from 86.05\% (based on clean data) to 48.07\%. We generated the CIFAR-10 dataset with TensorClog perturbations using default settings from their GitHub repository. Training the model in Table~\ref{fig:synarchitecture} on this dataset without our approach resulted in a test accuracy of 83.53\%. However, with our approach, the model's test accuracy notably increased to 90.01\%. These findings highlight the effectiveness of our approach in handling unlearnable data with TensorClog perturbations. However, OPA also resulted in a model with almost similar test accuracy of 88.05\%, showing that TensorClog perturbations are vulnerable to linear transformations.

\textbf{PUE~\cite{wang2024provably}:}
We generated Provably Unlearnable Examples (PUE) using the code from their GitHub repository. After training Table~\ref{fig:separchitecture}without applying nonlinear transformations, we achieved a test accuracy of 52.36\%. After applying nonlinear transformations, we improved the accuracy to 85.89\%. However, when using the linear transformation method (OPA), we achieved a higher accuracy of 90.09\%.

\subsection{Comparison with Adversarial Training}
To demonstrate our approach's effectiveness, we compared it with adversarial training, a prominent defense mechanism~\cite{DBLP:conf/iclr/FuHLST22, DBLP:conf/nips/Tao0WYHC22}. Following \cite{DBLP:conf/iclr/FuHLST22}'s approach, we conducted adversarial training using PGD attack~\cite{DBLP:conf/iclr/MadryMSTV18} with VGG19 and VGG16 as base models, ensuring compatibility with our architectures. Perturbation radius and step-size were set to 4/255 and 0.8/255 (default), respectively, for all unlearnable noises, except error-minimizing noise. For the latter, a perturbation radius of 8/255 and step-size of 2/255 were used, resulting in better accuracy than a perturbation radius of 4/255. Default values were maintained for other parameters, such as 10 PGD steps and 40000 training iterations. Test accuracies under adversarial training are reported in Column 7 in Table~\ref{t:5}, showcasing our approach's superiority across all considered noises.

\subsection{Perturbed 50\% of the Training Set}
In our experiments, we also consider the case in which only 50\% of the dataset is unlearnable. Similar to the experiment in Section~\ref{deepconfuse}, we performed on the CIFAR-10 crafted by Deepconfuse, error-minimizing, error-maximizing, synthetic, autoregressive, OPS, SEP, Entangled Features, REM, Hypocritical, and TensorClog approaches, respectively. We did not perform any nonlinear transformations for the training set and used similar models in Table \ref{fig:dcarchitectur}-\ref{fig:synarchitecture}. The models achieved a test accuracy of 85.65\%, 86.28\%, 86.49\%, 83.8\%, 87.56\%, 86.69\%, 87.81\%, 86.72\%, 86.73\%, 88.53\%, and 86.14\%, respectively. This also demonstrates the limited effectiveness of those data protection approaches. That is, these approaches are vulnerable to nonlinear transformations and ineffective when only half of the dataset is protected.

\subsection{Comparison with Existing Attack Methods}
\label{sec.compare}
In this section, we compare our proposed attack with existing attack methods, including AVATAR~\cite{dolatabadi2024devil}, CutMix~\cite{DBLP:conf/iccv/YunHCOYC19}, Mixup~\cite{zhang2017mixup}, and CutOut~\cite{devries2017improved} in terms of the test accuracy. AVATAR~\cite{dolatabadi2024devil} is a recently proposed attack method against unlearnable datasets, based on diffusion models. Moreover, CutMix~\cite{DBLP:conf/iccv/YunHCOYC19}, Mixup~\cite{zhang2017mixup}, and CutOut~\cite{devries2017improved} are advanced data augmentation techniques widely used to improve the generalization ability of DNN models. For implementation of AVATAR~\cite{dolatabadi2024devil}, we used the code from their GitHub repository~\cite{AVATAR-GitHub}, with a diffusion timestep of 100—the recommended setting for the CIFAR-10 dataset. All other parameters were kept at their default values. The implementations of CutMix, Mixup and CutOut were also sourced from the same repository.

The test accuracies for each attack method, including our proposed approach, are presented in Table~\ref{tab:comparison}. For most unlearnable datasets, AVATAR and our method achieve comparable performance, with AVATAR slightly outperforming ours in some cases. However, on autoregressive datasets, our attack method yields significantly higher test accuracy than all four baseline methods.

\begin{table*}[ht]
\centering
\caption{Our proposed method compared to existing attack methods and data augmentation techniques against 13 unlearnable datasets. The highest accuracy achieved for each dataset is highlighted in bold.}
\label{tab:comparison}
\renewcommand{\arraystretch}{0.5}
\begin{tabular}{|>{\color{black}}p{4.1cm}|>{\color{black}}p{1.2cm}|>{\color{black}}p{1.6cm}|>{\color{black}}p{1.4cm}|>{\color{black}}p{1.4cm}|>{\color{black}}p{1.4cm}|}
\hline
Data protection approach & Ours (\%) & AVATAR (\%) & CutMix (\%) & MixUp (\%) & CutOut (\%) \\
\hline
NTGA & 88.74 &	\textbf{88.83}	& 12.73 & 13.32	& 12.03 \\
\hline
Deepconfuse & 87.00	& \textbf{90.29} &	17.31 & 26.65	& 14.86 \\
\hline
Error-minimizing & 85.2 &	\textbf{86.68} &	21.48  & 33.71	& 20.39 \\
\hline
Error-maximizing & \textbf{92.2}	& 91.92	& 30.31 & 53.28	& 28.3\\
\hline
Synthetic & 88.2	& \textbf{91.75} &	16.28 & 25.31	& 16.56\\
\hline
Autoregressive & \textbf{86.9}	& 36.00 &	10.63 & 12.27 &	11.28\\
\hline
OPS & 86.71	& \textbf{89.29} & 79.37  & 33.19	& 56.55 \\
\hline
SEP & 88.18	& \textbf{92.18}&	18.11  & 44.84	& 16.93 \\
\hline
EntF & 88.59	& 89.95	& \textbf{92.85}  & 93.36 &	92.75 \\
\hline
REM & 86.21&	\textbf{91.06}	& 20.5 & 15.94	& 22.67 \\
\hline
Hypocritical & \textbf{89.68}	& 91.65	& 16.88 & 18.04	& 15.44 \\
\hline
TensorClog & 88.84	& 91.78& 87.39 & \textbf{92.77} & 91.85 \\
\hline
PUE & 85.4	& \textbf{91.83}& 8.83 &  11.72 & 10.15\\
\hline
\end{tabular}

\end{table*}

\begin{figure}
    \centering
    \includegraphics[width=0.85\linewidth]{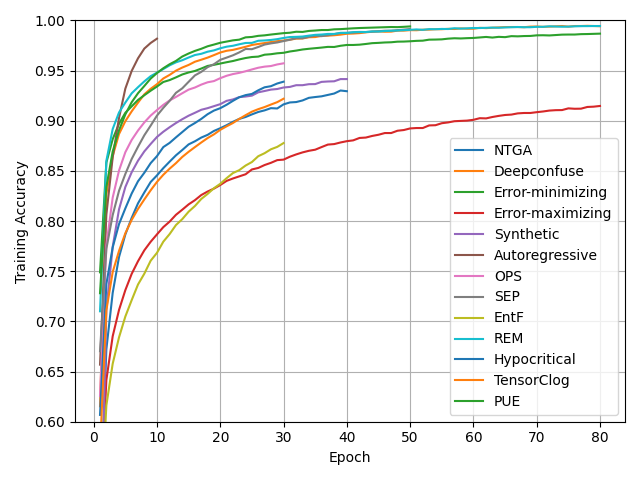}
    \caption{Training accuracies of models trained with unlearnable CIFAR-10 datasets.}
    \label{fig:training_acc}
\end{figure}

\section{Discussion} \label{discussion}
Section~\ref{sec:evaluation} provides an evaluation of 13 data protection approaches, revealing their vulnerability to nonlinear transformations and the subsequent degradation of protection levels. Fig.~\ref{fig:training_acc} shows the training accuracies of the models trained using our approach. Some models achieve over 95\% accuracy, yet their test accuracies given in Column 4 in Table~\ref{t:5}) remain lower, indicating some some degree of overfitting for specific unlearnable datasets. The length of the curve represents the number of epochs each model was trained. We carefully determined the number of epochs by incrementally increasing them (e.g., 10, 20) in each experiment until reaching the target test accuracy. Models trained on certain datasets—specifically DeepConfuse, Error-minimizing, Error-maximizing, and REM—required more training epochs than others to achieve the target accuracy. This suggests that our approach facilitates the transformation of unlearnable data into learnable, albeit with potential disparities in the distributions of training and test data. 

In this section, we discuss additional experimental results. To evaluate the effectiveness of our approach on diverse datasets, we experimented on unlearnable MNIST and ImageNet datasets crafted by the NTGA. We also present an experimental evaluation for an additional data protection approach, CUDA \cite{sadasivan2023cuda}. Recent studies in \cite{liu2023image, qin2023learning, DBLP:journals/corr/abs-2305-19254} have delved into approaches for breaking unlearnable datasets. We provide a comprehensive discussion outlining the distinctions between our approach and theirs.

\textbf{NTGA on MNIST:} As shown in~\cite{yuan2021neural}, the lowest test accuracy of the model trained based on unlearnable MNIST dataset is around 16\% for Convolutional Neural Networks (CNNs). Our objective is to demonstrate that these so-called ``unlearnable data" can achieve a test accuracy of 98\%, matching the performance of a model trained on clean data. In essence, we elevate the test accuracy from 16\% to 98\% through image transformation techniques outlined in Section~\ref{Appendix:DataAug}. Initially, we employed image nonlinear transformation methods to render unlearnable data learnable, utilizing the Keras ImageDataGenerator function with attributes like a rotation range of 10 and a zoom range of 0.1. Subsequently, we applied the threshold binary transformation with a threshold value of 0.5 and a maximum value of 1 to the training dataset. Lastly, we utilized the JPEG Compression transformation. Table~\ref{fig:modelmnist} shows the model used for this experiment. The model consists of two convolution layers with ReLU activation function and two fully connected (FC) layers. The first FC layer has 1024 units with an ReLU activation function, and the next one has ten units with the softmax activation function. In addition, a dropout layer is added between the two FC layers that randomly drops 25\% of the weights. The model was trained for ten epochs with a batch size of 100. This setup gave a test accuracy of 98.64\%, the same as the accuracy obtained based on clean data. In contrast, employing the same model architecture without nonlinear transformations resulted in a mere test accuracy of 17.50\%. This showcases the vulnerability of the unlearnable MNIST dataset crafted by NTGA to the effects of data augmentation. 

\textbf{NTGA on ImageNet:} The test accuracy remains around 70\% for most model architectures in Yuan and Wu's study on their unlearnable ImageNet dataset~\cite{yuan2021neural}. We performed the proposed framework using the Keras ImageDataGenerator by setting the rotation range to 2, the horizontal flip to True, and the zoom range to 0.1. Moreover, the training dataset was increased up to three times the original dataset size using nonlinear transformations in the OpenCV package. These nonlinear transformations are color channel manipulation, thresh binary, and thresh binary inverse, with a threshold value of 0.5 and a maximum value of 1.

The baseline model we used for this experiment is ResNet with 152 convolution layers (ResNet152) and random initialization of weights. We extended the model by adding six FC layers and one dropout layer after a series of convolutional layers from ResNet152. These FC layers have 1024, 512, 256, 128, and 64 units with an ReLU activation function. The last layer has two neurons with a softmax activation function. Furthermore, we added a dropout layer after the first layer, which will lead to a 30\% random drop of the model weights. Table \ref{fig:ImageNet model} provides the details of the specifications of the model architecture. We trained the model for 100 epochs with a batch size of 10 and a learning rate of 0.001. This setup yielded a training accuracy of 96.81\%, and a test accuracy of 94.28\%. The test accuracy closely matches models trained on clean data, underscoring the vulnerability of NTGA to our approach. Employing the same model architecture without our framework yielded a test accuracy of only 75.71\%. This outcome underscores the significant impact of nonlinear transformations on the success or failure of data protection approaches, such as NTGA in the above experiments.

\textbf{CUDA~\cite{sadasivan2023cuda}:} As proposed by~\cite{sadasivan2023cuda}, we generated a Convolution-based Unlearnable Dataset (CUDA) using the code provided in their Git-Hub repository. We used a blur parameter of 0.3 to obtain a dataset with enhanced protection. Initially, when a VGG16 model was trained on these images, the test accuracy was only 10.56\%. However, by expanding the training dataset through cropping and dilating techniques and incorporating the Keras ImageDataGenerator, we achieved a significantly improved test accuracy of 43.35\%. Unlike other approaches, such as error-minimizing, error-maximizing, and NTGA, CUDA is not model-dependent and does not produce additive noises. CUDA introduces multiplicative noise, resulting in more noise in the image's background. As most of the noise in this dataset is in the background, we noticed that cropping the background is effective.

Moreover, we applied a series of transformations to the CUDA dataset. Initially, we cropped 1 pixel from each side of the borders. Then, we implemented a horizontal flip with a probability of 0.5. Contrast and brightness were increased by 50\% with a probability of 0.7. Next, we increased saturation and sharpness to twice their existing values with a probability of 0.7. We further added a contour filter with a probability of 0.2 and increased the hue channel by 10. For images that did not have a contour filter added, we used a posterize filter with a probability of 0.7. Finally, we applied grayscale with a probability of 0.2. Training these transformed unlearnable images on the model shown in Table~\ref{fig:emaxarchitectur} without the first fully connected layer with 2048 neurons achieved a test accuracy of 77.36\%. However, the model's training accuracy is 81.22\%. Additionally, further exploration of more advanced and specified transformations is necessary to break the CUDA-protected dataset, especially considering CUDA's significant difference in methodology from other approaches.

\textbf{Unlearnable Clusters~\cite{zhang2023unlearnable}:} We generated the unlearnable Pets~\cite{parkhi2012cats} dataset using the approach in~\cite{zhang2023unlearnable}. We utilized the code from their GitHub repository. We downloaded the clean Pets dataset from the PyTorch datasets. The training dataset consists of 3,680 images belonging to 37 classes, and the test dataset contains 3,669 images. Following the  settings given in~\cite{zhang2023unlearnable}, we employed ResNet50 as the surrogate model and used 10 clusters to produce the perturbations. In our experiments, we used a pretrained ResNet50 with ImageNet weights, a global average pooling layer, and a fully connected layer with 37 neurons with a softmax activation function corresponding to the 37 classes. First, we trained the model on the clean Pets dataset and achieved a test accuracy of 84.98\%. When we trained the same model on the unlearnable Pets dataset generated by~\cite{zhang2023unlearnable}, the test accuracy dropped to 46.39\%. Therefore, our goal was to attack the unlearnable Pets dataset by improving the test accuracy. After applying our approach, we were able to achieve a model with a test accuracy of 73.54\% . We used the pixel manipulation, erode, and dilate transformations to overcome the effect of unlearnable perturbations generated by Zhang et al.~\cite{parkhi2012cats}. Hence, we can conclude that our proposed approach can attack the unlearnable data generated by Zhang et al.~\cite{parkhi2012cats}.

\textbf{UEraser~\cite{qin2023learning}} is a recently proposed approach to break unlearnable data. In contrast to our approach, they utilized modern image transformation techniques such as PlasmaTransform and ChannelShuffle. Initially, we replicated their experiments using the code in their GitHub repository. UErasor is applied on the unlearnable CIFAR-10 dataset with synthetic perturbation, which resulted in a test accuracy of 91.97\%. We executed the code without UErasor, and the test accuracy is 19.95\%. Then, we replaced the modern image transformation in UEraser with the nonlinear transformation techniques we used. They are rotate, resize, flip, brightness, and grayscale. This modification led to a test accuracy of 88.97\%. This demonstrates that despite their use of modern image transformation techniques, our approach remains almost as effective as UErasor.

Furthermore, the following work is also related to our research.

\textbf{Image Shortcut Squeezing (ISS)~\cite{liu2023image}} explored an attack method against unlearnable data based on simple compression techniques. They mainly used grayscale and compression methods, such as JPEG compression, to mitigate the effect of unlearnability. Their main focus was on evaluating the effectiveness of image squeezing methods against unlearnable data, while our study concentrates on a broader area, including nonlinear transformations and building a framework to overcome unlearnability. ISS was able to improve CIFAR-10 model accuracy to 81.73\% for  12 existing unlearnable methods. We successfully assessed the same eleven approaches, and achieved a test accuracy exceeding 85\%, including the EntF approach~\cite{DBLP:conf/iclr/00020000023}, which they had not explored. The ShortcutGen dataset~\cite{DBLP:journals/corr/abs-2211-01086} is the only approach we did not consider because the code is not publicly available.

\section{Conclusion} \label{sec:conclusion}

Recent advancements in defense mechanisms, such as NTGA and Deepconfuse, aim to safeguard data against unauthorized deep learning use. However, our research exposed vulnerabilities in these approaches, particularly when confronted by the proposed nonlinear transformation framework. Testing on CIFAR-10, ImageNet, and MNIST datasets revealed that data assumed to be unlearnable could achieve over 85\% accuracy through our proposed nonlinear transformation techniques, compromising the efficacy of existing data protection measures. Our approach provides a model with improved test accuracy than the existing linear separable approach given in~\cite{DBLP:journals/corr/abs-2305-19254}  on eleven CIFAR10 datasets. This highlights a significant gap in current defense methods, as even partial clean datasets exhibit high accuracy. Our findings underscore the importance of exploring more effective data protection approaches and developing robust data protection approaches capable of withstanding such techniques. This emphasizes the necessity of considering nonlinear transformations in the future development of resilient data protection approaches.


\bibliographystyle{cas-model2-names}
\bibliography{sample-base}









\end{document}